\RequirePackage{amsmath}
\documentclass[runningheads]{llncs}

\usepackage{amsmath}
\usepackage[utf8]{inputenc}
\usepackage{authblk}
\usepackage{color,soul}
\usepackage[dvipsnames]{xcolor}
\usepackage{mathtools ,amssymb ,amsthm}
\usepackage{graphicx}
\usepackage{tabularx,ragged2e}
\usepackage{hyperref}
\usepackage{makecell}
\usepackage{float}
\usepackage{verbatim}
\usepackage{adjustbox}
\usepackage{multirow}
\usepackage{booktabs}
\usepackage{algorithm} 
\usepackage{algpseudocode}
\usepackage{subcaption}
\usepackage[T1]{fontenc}
\usepackage{tikz}
\usepackage{longtable}
\usepackage{array}
\usepackage{pdflscape}

\DeclareUnicodeCharacter{2212}{-}

\begin{document}
\title{Explainable Detection of Implicit Influential Patterns in Conversations via Data Augmentation}
\titlerunning{Detection of Implicit Influential Patterns in Conversations}
%
\author{Sina Abdidizaji\inst{1} \and
Md Kowsher\inst{1} \and
Niloofar Yousefi\inst{1} \and
Ivan Garibay\inst{1}}
\authorrunning{S. Abdidizaji et al.}
%
\institute{Industrial Engineering and Management Systems, University of Central Florida, Orlando, FL, USA \\
\email{\{sina.abdidizaji, md.kowsher, niloofar.yousefi, igaribay\}@ucf.edu}}

\maketitle              
\begin{abstract}
In the era of digitalization, as individuals increasingly rely on digital platforms for communication and news consumption, various actors employ linguistic strategies to influence public perception. While models have become proficient at detecting explicit patterns, which typically appear in texts as single remarks referred to as utterances, such as social media posts, malicious actors have shifted toward utilizing implicit influential verbal patterns embedded within conversations. These verbal patterns aim to mentally penetrate the victim's mind in order to influence them, enabling the actor to obtain the desired information through implicit means. This paper presents an improved approach for detecting such implicit influential patterns. Furthermore, the proposed model is capable of identifying the specific locations of these influential elements within a conversation. To achieve this, the existing dataset was augmented using the reasoning capabilities of state-of-the-art language models. Our designed framework resulted in a 6\% improvement in the detection of implicit influential patterns in conversations. Moreover, this approach improved the multi-label classification tasks related to both the techniques used for influence and the vulnerability of victims by 33\% and 43\%, respectively.
\end{abstract}

\section{Introduction}
In the era of computers and digital interactions, individuals are increasingly exposed to risks they do not anticipate or desire. Social media, instant messaging applications, and chatbots serve as digital platforms that enable individuals to interact without being physically present or revealing their identity and face. To enhance safety and ensure secure and healthy digital platforms for such interactions, it is crucial to effectively identify influential statements and behaviors. A secure and healthy digital environment enables appropriate interactions without threats from malicious actors seeking to influence users in order to steal information, issue threats, or endanger individuals for their own objectives. On these platforms, where all communication occurs digitally, there are heightened opportunities for phishing, scamming, and other manipulative actions designed to extract sensitive information from users \cite{zaidyCybersecurityPersonalPrivacy2024}. Addressing these challenges and detecting such behaviors on digital platforms and chatbots are essential for maintaining a secure environment for interactions via computers and other digital devices. As current models have achieved high accuracy in detecting explicit patterns, malevolent actors are increasingly attempting to exert mental influence over users to accomplish their aims. Mental influential patterns constitute deceptive strategies intended to control or influence the emotions, thoughts, and behaviors of targeted individuals \cite{barnhillWhatManipulation2014,guoUnderstandingCombatingOnline2023}. It represents an intersection of mental health conditions and toxic behavior, characterized by causing distress through implicitly deceitful remarks \cite{wangMentalManipDatasetFinegrained2024}. Unlike explicit hate speech or overtly toxic language, influential statements are inherently subtle, nuanced, and difficult to detect. Recently, actors have increasingly used nuanced strategies to influence audiences through conversational contexts. Detecting such remarks has proven to be significantly more challenging than identifying hate speech \cite{elsheriefLatentHatredBenchmark2021,ghoshCoSynDetectingImplicit2023}, toxicity \cite{atwellAPPDIADiscourseawareTransformerbased2022}, or sarcasm \cite{akulaInterpretableMultiHeadSelfAttention2021}. Previous detection models typically relied on learning from labeled sentences or paragraphs. However, current influential remarks often manifest subtly within broader conversations, appearing sporadically in sentences \cite{ziemsCanLargeLanguage2024}. This intermittent nature complicates the detection task for language models. Moreover, mental influential patterns often lack overtly negative connotations, becoming identifiable only when analyzed within the context of an entire conversation.

The objective of this research is to enhance the detection of implicit influential patterns within conversations using the capabilities of large language models. Previous studies \cite{maDetectingConversationalMental2024,yangEnhancedDetectionConversational2024} have shown that relying solely on prompting with available large language models is not an effective approach for detecting such patterns. Even fine-tuning models on conversations with a single label has not proven to be as effective as anticipated \cite{wangMentalManipDatasetFinegrained2024}. To address this gap, we propose a framework designed to improve the accuracy of detection tasks. This framework consists of two main stages: data augmentation and a two-phase fine-tuning process. Specifically, our augmentation strategy involves utilizing a reasoning language model to identify mental influential statements within conversations. The detected influential sentences are subsequently incorporated into the fine-tuning pipeline, in order to boost overall model performance. Another motivation for this augmentation strategy is to improve model interpretability through instruction fine-tuning \cite{weiFinetunedLanguageModels2021}. By training the model to precisely identify the locations of mental influential elements within conversations, we can develop an explanatory system capable of highlighting and clarifying these influential segments in a conversation. 

The structure of this paper is organized as follows. The following section reviews related work and relevant literature. Section 3 provides a detailed explanation of the proposed framework. Section 4 describes the datasets and experimental setup, while Section 5 presents the results. Finally, Section 6 concludes the paper.

\section{Related Work}
There are numerous studies examining influence both in general and specifically within texts. \cite{abdidizajiAnalyzingXsWeb2025} investigated influential actors on the X social media platform by analyzing the frequency of news sharing, finding that individuals who share news with varying credibility and platform popularity exhibit distinct influence patterns across the network. In the context of textual analysis, \cite{ziemsCanLargeLanguage2024} categorize text data into three groups: utterances, conversations, and documents. An utterance typically refers to a standalone statement produced by an individual, such as a post on an online social networking platform, and does not require conversational engagement \cite{bakhtinSpeechGenresOther1986}. Notably, an utterance may consist of several sentences.

Several datasets focus on utterances collected from online forums and social networks. For instance, Dreaddit \cite{turcanDreadditRedditDataset2019} addresses mental stress, and Detex \cite{yavnyiDeTexDBenchmarkDataset2023a} focuses on delicate text. As utterances can be produced by large language models, datasets such as ToxiGen \cite{hartvigsenToxiGenLargeScaleMachineGenerated2022} have been generated using these models to provide numerous training samples aimed at enhancing safety and mitigating hate speech. While progress in utterance-level detection has been significant, more sophisticated models are required for conversation and document-level tasks \cite{ziemsCanLargeLanguage2024}. Within the field of human-computer interaction, social chatbots have been developed to help users cope with mental distress. However, a recent study \cite{laestadiusTooHumanNot2024} reported that prolonged communication with such chatbots can result in mental health harms, primarily due to users’ emotional dependence on these systems, which develops over the course of continuous interactions between the individual and the computer.

Recent research has aimed to improve the detection of influential patterns by employing advanced prompting methods, such as Chain-of-Thought (CoT) \cite{yangEnhancedDetectionConversational2024} and intent-aware prompting techniques \cite{maDetectingConversationalMental2024}. Incorporating Chain-of-Thought prompts \cite{kojimaLargeLanguageModels} for detecting implicit influential patterns did not significantly improve results, although a combination of CoT with few-shot learning yielded modest gains \cite{yangEnhancedDetectionConversational2024}. Intent-aware prompting involves first extracting the intent of each participant in a conversation using a language model, then appending this information to the conversation and prompting the model again to detect mental manipulation. This approach demonstrated greater improvement in detection performance compared to other methods \cite{maDetectingConversationalMental2024}. A recent study \cite{gaoMentalMACEnhancingLarge2025} introduced MentalMAC, a multi-task anti-curriculum distillation approach for mental manipulation detection. By leveraging a large teacher model to generate rationales and feedback, they combined unsupervised data augmentation (EVOSA) with staged knowledge distillation to train a smaller student model. Their student model surpassed larger LLMs, achieving higher accuracy than established baselines.

As these studies demonstrate, many methods involve augmenting conversational data by adding information extracted from the primary data source. To further improve detection accuracy, we propose a novel framework for detecting implicit influential patterns in conversations, featuring new data augmentation and fine-tuning approaches.

\section{Methodology}
In this section, the designed framework for implicit influential patterns detection will be explained extensively. First, data augmentation will be explained and then we leverage the augmented data to fine-tune a base language model in two phases for having a robust model.

\subsection{Data Augmentation}
\begin{figure}[H]
    \centering
    \includegraphics[width=0.9\textwidth]{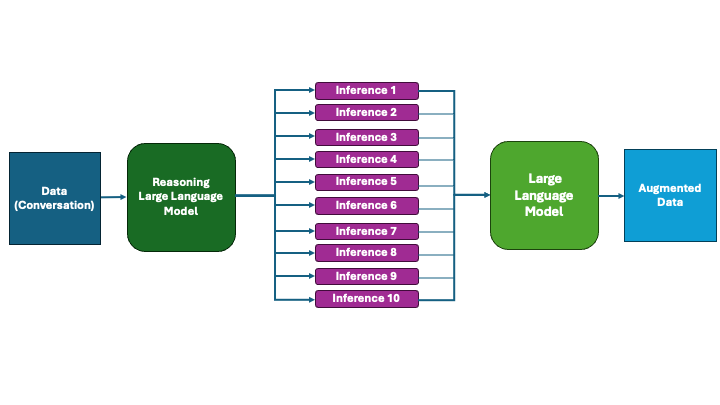}
    \caption{Data augmentation pipeline for finding influential patterns}
    \label{fig1}
\end{figure}

In the proposed framework, instead of training the model on the entire conversation and providing a single label, the objective is to indicate which parts of the conversation contain implicit influential patterns manifested as mental manipulation. The conversations are between two individuals and are separated line by line. Reasoning language models are leveraged to identify the specific lines that contain implicit influential elements. Through this approach, the augmented data provides the model with the particular lines that need to be learned to better detect influential parts, rather than presenting the whole conversation with a single binary label. To accomplish this, distilled versions of the Deepseek language model \cite{deepseek-aiDeepSeekR1IncentivizingReasoning2025a} — which are open source and available online, particularly the Llama-distilled variant — were employed to identify influential segments. Given the stochastic nature of these models, each conversation was prompted to the reasoning language model ten times, and the results from these analyses were summarized by another language model. Notably, the summarization is performed by a language model that does not conduct reasoning. Further details are provided in \hyperref[app:appendix1]{Appendix~1}. The detailed pipeline for data augmentation is presented in \autoref{fig1}.

After identifying these influential segments within conversations, we manually sampled the results to verify the accuracy of this approach. Since each conversation was independently analyzed ten times and the results were aggregated, the data augmentation process demonstrated high accuracy.

\subsection{Model Framework}
The primary rationale for data augmentation is to train the model to identify the locations of influential statements within conversations, thereby enhancing its learning capacity. Given the computational expense of fully fine-tuning language models, instruction fine-tuning is employed by attaching a Low-Rank Adapter (LoRA) \cite{huLoRALowRankAdaptation2021a} to the model. This instruction-tuned model is then used to identify implicit influential segments within conversations, a task previously unattainable due to the lack of relevant data.

LoRA introduces an approach in which, instead of fully fine-tuning all layers in a neural network, the weight updates are approximated by two low-rank matrices, which are then attached to the layers. This approach is also advantageous because all the base model weights can be frozen, allowing only the newly added parameters introduced by the low rank adapters to be trained \cite{huLoRALowRankAdaptation2021a}. Mathematically, if the initial weights are represented by a matrix $W_1 \in \mathbb{R}^{d \times k}$, the weight updates can be approximated by two matrices, $A \in \mathbb{R}^{d \times r}$ and $B \in \mathbb{R}^{r \times k}$, where the rank $r$ should be chosen such that $r < \min(d, k)$, where $d$ is the input dimension and $k$ is the output dimension, respectively. Thus, the weight matrix for the instruction fine-tuned model is given by:
\begin{equation}
    h_1 = W_1 x + \Delta W_1 x = W_1 x + ABx
    \label{eq:lora_instruction}
\end{equation}
where $h_1$ represents the forward pass of the instruction fine-tuned model, $W_1$ denotes the initial weights of the language model, $\Delta W_1$ represents the weight updates from instruction fine-tuning, $x$ denotes the concatenation of the instruction prompt and the initial conversation as input data, and the labels correspond to the augmented data generated in the previous procedure.

\begin{figure}[htbp]
    \centering
    \includegraphics[width=0.9\textwidth]{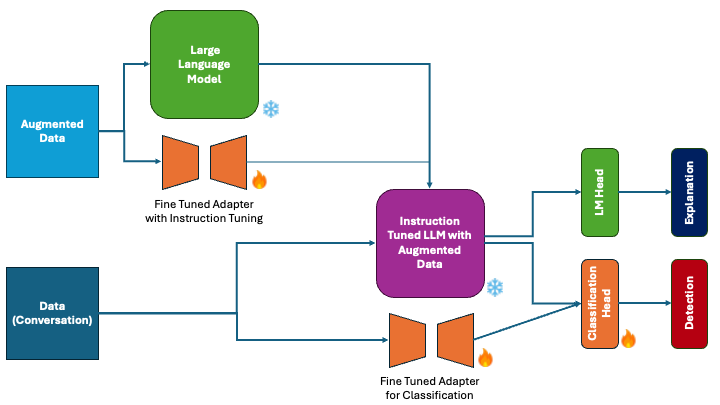}
    \caption{The framework of two-phase fine-tuning for detecting mental influential patterns. The snowflake symbol indicates frozen weights, whereas the fire symbol denotes the weights that are updated during the fine-tuning process.}
    \label{fig2}
\end{figure}

For classification tasks, referred to as detection in our framework, the newly attached adapter and the new classification head — added after removing the original language model head — are simultaneously fine-tuned. Following the initial instruction fine-tuning, the previous weights are frozen, and only the newly introduced parameters in the second adapter and the classification head are updated. Prior to removing the language model head and attaching the classification head, as a new adapter is added to the instruction fine-tuned model, this step can be mathematically expressed as:
\begin{equation}
    h_2 = W_2 y + \Delta W_2 y = (W_1 + \Delta W_1) y + C D y = W_1 y + A B y + C D y
    \label{eq:lora_classification}
\end{equation}
where $h_2$ denotes the forward pass of the model with the newly attached adapter, $W_2$ represents the initial weights of the language model after instruction fine-tuning, including the weights from the adapter attached during the first stage, $\Delta W_2$ denotes the weight updates from classification training, $y$ is the original conversation input for the classification task, and $C$ and $D$ are analogous to the $A$ and $B$ matrices but may have different dimensions. The attached classifier is then trained to determine whether or not a conversation contains implicit influential patterns. It should be noted that open-source models from the Llama 3 series \cite{grattafioriLlama3Herd2024} are used in the experiments, as they can be downloaded and fine-tuned specifically for our tasks \footnote{Code: \url{https://github.com/sina6990/IMM}}. The complete model framework is illustrated in \autoref{fig2}.

\section{Experimental Setup}
\subsection{Dataset}
Most publicly available datasets are based on individual utterances. We have excluded these datasets and instead focused on newly released datasets compiled by \cite{wangMentalManipDatasetFinegrained2024}. These datasets comprises approximately 4,000 conversations and includes three distinct types of labels: the presence or absence of mental influence, multi-label annotations specifying the techniques employed for mental influence, and the vulnerability types of influenced victims. Our goal is to improve detection accuracy across all these categories. This paper \cite{wangMentalManipDatasetFinegrained2024} introduces two datasets: MentalManipCon and MentalManipMaj. In these names, "con" stands for consensus, while "maj" denotes majority. During the annotation process, annotators sometimes held differing opinions. The consensus dataset contains labels assigned only when all annotators were in agreement, whereas the majority dataset includes labels determined by the majority vote among annotators, even in the presence of differing viewpoints. For further details regarding the annotation procedure, read the original paper by \cite{wangMentalManipDatasetFinegrained2024}. 

The technique labels used for identifying techniques of mental influence are: "Denial", "Evasion", "Feigning Innocence", "Rationalization", "Playing the Victim Role", "Playing the Servant Role", "Shaming or Belittlement", "Intimidation", "Brandishing Anger", "Accusation", and "Persuasion or Seduction". The vulnerability labels for victims are: "Over-responsibility", "Over-intellectualization", "Naivete", "Low self-esteem", and "Dependency". For further details and definitions of each technique and vulnerability label, refer to the paper by \cite{wangMentalManipDatasetFinegrained2024}.

\subsection{Evaluation Metrics}
The experiments are divided into two parts. The first part involves binary classification, where the trained model predicts whether a given conversation contains any implicit influential patterns. The second part involves multi-label classification, in which the model is required to identify all relevant technique labels used by the actors influencing the victims, as well as the vulnerability labels of victims present in a conversation. The primary evaluation criterion is accuracy, along with other standard metrics such as precision, recall, and micro F1 score \cite{jamesIntroductionStatisticalLearning2023}.

\section{Results}
\subsection{Binary Classification of Implicit Influential Patterns}
First, the detection of influential patterns was investigated using zero-shot and few-shot learning approaches with state-of-the-art large language models. Zero-shot learning \cite{larochelleZerodataLearningNew2008} refers to querying a vanilla language model without any additional training or fine-tuning, assessing its performance based solely on the knowledge acquired during pretraining and post-training phases. In other words, the model is evaluated based on the general knowledge it has acquired during training phases, without having been explicitly trained on the specific task being assessed. As shown in \autoref{tab1}, zero-shot learning did not yield significant performance differences across different models. However, the newer 3.2 version of the Llama model with 3 billion parameters outperformed other model variants. It is noteworthy that the smallest Llama model, with only 1 billion parameters, performed poorly in zero-shot learning, likely due to its limited capacity for storing knowledge.

Few-shot learning \cite{brownLanguageModelsAre2020b} is similar to zero-shot learning, except that a few labeled examples are included in the original prompt before querying the model. This approach tests whether the language model can identify conversations containing implicit influential patterns when given some guidance through examples. For the few-shot learning experiments, two positive and two negative examples were included in each prompt, with examples randomly selected from the dataset; thus, the prompts did not always contain the same samples. The results indicate that the largest model size achieved the best performance among all evaluated models. Notably, few-shot learning improved the performance of the smallest model by 34 percent. This finding suggests that, although the Llama-3.2-1B model alone lacked sufficient internal knowledge, providing relevant examples enabled it to better detect influential patterns compared to zero-shot learning with only the conversation itself.
As zero-shot and few-shot approaches yielded only limited improvements compared to previous iterations of such models, these results underscore the necessity of a robust pipeline to enhance detection accuracy, since larger models do not necessarily yield better results. Therefore, we conducted experiments using the proposed framework outlined in the methodology section, and the results are reported in \autoref{tab1} under “ours” alongside the baseline models. The highest accuracy was achieved by the model utilizing Llama-3.2-3B as the base language model, with an accuracy of 82.6 percent on the MentalManipCon dataset. The other two models performed comparably, resulting in an overall performance improvement of approximately 6 percent. Notably, this improvement was attained by fine-tuning a language model with 10 billion fewer parameters. Even when Llama-3.2-1B was used as the base model, the performance remained around 82 percent, utilizing 12 billion fewer parameters. This demonstrates that designing a robust fine-tuning pipeline is more critical than merely increasing model size or fine-tuning on raw data.

For the other dataset, MentalManipMaj, the Llama-3.1-8B model achieved the best results. The performance of Llama-3.2-3B was also noteworthy and comparable to Llama-3.1-8B, with both models improving accuracy by around 3 percent. Although Llama-3.2-1B did not achieve the same level of improvement as the larger models, it still outperformed the model trained with the approach from \cite{wangMentalManipDatasetFinegrained2024}, despite having 12 billion fewer parameters under this framework.
\begin{table}[htbp]
\centering
\caption{Performance of Llama models in terms of accuracy, precision, recall, and F1-score under zero-shot, few-shot, and fine-tuning settings}
\label{tab1}
\resizebox{\textwidth}{!}{
\begin{tabular}{c cccc c cccc c} 
\hline
\multirow{2}{*}{\textbf{Model}}
    & \multicolumn{4}{c}{\textbf{MentalMalipCon Dataset}}
    & \multicolumn{1}{c}{} 
    & \multicolumn{4}{c}{\textbf{MentalMalipMaj Dataset}}
    & \multirow{2}{*}{\textbf{Reference}} \\
\cline{2-5} \cline{7-10}
 & {\fontsize{7}{10}\selectfont \textbf{Accuracy}} & {\fontsize{7}{10}\selectfont \textbf{Precision}} & {\fontsize{7}{10}\selectfont \textbf{Recall}} & {\fontsize{7}{10}\selectfont \textbf{F1}}
 & 
 & {\fontsize{7}{10}\selectfont \textbf{Accuracy}} & {\fontsize{7}{10}\selectfont \textbf{Precision}} & {\fontsize{7}{10}\selectfont \textbf{Recall}} & {\fontsize{7}{10}\selectfont \textbf{F1}}
 & \\ \hline

\multicolumn{11}{c}{\textbf{Zero-shot Learning}} \\ \hline
Llama-2-13B   & .696 & .693 & \textbf{.997} & .696 & & .721 & .722 & \textbf{.997} & .721 & \cite{wangMentalManipDatasetFinegrained2024} \\
Llama-3.1-8B  & .695 & .696 & .993 & .695 & & .707 & .708 & .993 & .707 & - \\ 
Llama-3.2-3B  & .705 & .710 & .970 & .705 & & .711 & .719 & .966 & .711 & - \\
Llama-3.2-1B  & .330 & .774 & .044 & .330 & & .319 & .785 & .045 & .319 & - \\ \hline

\multicolumn{11}{c}{\textbf{Few-shot Learning}} \\ \hline
Llama-2-13B   & .715 & .735 & .912 & .715 & & .726 & .732 & .979 & .726 & \cite{wangMentalManipDatasetFinegrained2024} \\
Llama-3.1-8B  & .696 & .701 & .978 & .696 & & .704 & .721 & .945 & .704 & - \\
Llama-3.2-3B  & .694 & .716 & .922 & .694 & & .707 & .716 & .968 & .707 & - \\
Llama-3.2-1B  & .677 & .707 & .910 & .677 & & .596 & .729 & .679 & .596 & - \\ \hline

\multicolumn{11}{c}{\textbf{Fine-tuning}} \\ \hline
Llama-2-13B   & .768 & .828 & .835 & .768 & & .748 & \textbf{.809} & .851 & .748 & \cite{wangMentalManipDatasetFinegrained2024} \\
Llama-3.1-8B (ours)  & .817 & .829 & .932 & .877 & & \textbf{.786} & .797 & .938 & \textbf{.862} & - \\
Llama-3.2-3B (ours) & \textbf{.826} & \textbf{.871} & .890 & \textbf{.881} & & .781 & .799 & .924 & .857 & - \\
Llama-3.2-1B (ours) & .819 & .853 & .911 & .880 & & .765 & .785 & .921 & .848 & - \\ \hline

\end{tabular}
}
\end{table}
\subsection{Multi-label Classification of Techniques and Vulnerabilities}
In \autoref{tab2}, the results for multi-label classification are provided. There were 11 unique techniques in total, and some may have one of them and others may have multiple techniques annotated for a manipulative conversation. Since the performance of a model needed to be tested to see how many of those labels can be detected for each conversation, a multi-label classification was required. Our method with a Llama base model with 8 billion parameters was the best one among the others, with the accuracy of 35.7 percent. Due to having a lot of different labels, a larger model performed better. It should be noted that our approach with the smallest Llama model with 1 billion parameters acheived a performance more than 10 times better than the vanilla fine-tuning in \cite{wangMentalManipDatasetFinegrained2024}. For vulnerability, since it has only 5 unique labels, the performance expected to be better due to lower complexity. In terms of accuracy, the performance of our approach with Llama base model with 3 billion parameters was the best among others. Nevertheless, the performance of our method with Llama-8B was on par with 3 billion parameter. The results clearly shows that have the approach was clearly a better option than using vanilla fine-tuning with a model with a lot of parameters and this can reduce costs in terms of hardware and boost acceleration since running model with a lot of parameters need high computation resources.
\begin{table}[htbp]
\centering
\caption{Performance of Llama models in terms of accuracy, precision, recall, and F1-score for multi-label classification of techniques used by manipulators and vulnerability of victims under fine-tuning settings}
\label{tab2}
\resizebox{\textwidth}{!}{
\begin{tabular}{c cccc c cccc c} 
\hline
\multirow{2}{*}{\textbf{Model}}
    & \multicolumn{4}{c}{\textbf{Technique}}
    & \multicolumn{1}{c}{} 
    & \multicolumn{4}{c}{\textbf{Vulnerability}}
    & \multirow{2}{*}{\textbf{Reference}} \\
\cline{2-5} \cline{7-10}
 & {\fontsize{7}{10}\selectfont \textbf{Accuracy}} & {\fontsize{7}{10}\selectfont \textbf{Precision}} & {\fontsize{7}{10}\selectfont \textbf{Recall}} & {\fontsize{7}{10}\selectfont \textbf{F1}}
 & 
 & {\fontsize{7}{10}\selectfont \textbf{Accuracy}} & {\fontsize{7}{10}\selectfont \textbf{Precision}} & {\fontsize{7}{10}\selectfont \textbf{Recall}} & {\fontsize{7}{10}\selectfont \textbf{F1}}
 & \\ \hline

Llama-2-13B   & .029 & .349 & \textbf{.821} & .490 & & .008 & .265 & \textbf{.756} & .393 & \cite{wangMentalManipDatasetFinegrained2024} \\
Llama-3.1-8B (ours)  & \textbf{.357} & \textbf{.569} & .529 & \textbf{.529} & & .438 & .512 & .488 & .493 & - \\
Llama-3.2-3B (ours) & .345 & .536 & .473 & .488 & & \textbf{.446} & \textbf{.549} & .541 & \textbf{.534} & - \\
Llama-3.2-1B (ours)  & .317 & .536 & .474 & .488 & & .405 & .483 & .450 & .460 & - \\ \hline

\end{tabular}
}
\end{table}

\section{Conclusion}
In this paper, a new framework was introduced to enhance the detection of implicit influential patterns that manifest as mental manipulation in conversations. Leveraging the reasoning abilities of large language models, conversations were augmented with lines containing influential patterns, and a base language model was subsequently trained using a two-stage fine-tuning pipeline to improve detection accuracy. Using this framework, the binary classification task, determining whether a conversation contains implicit influential patterns, was improved by 6 percent. Multi-label classification tasks for detecting the techniques used and the vulnerability of victims were improved by 33 percent and 43 percent, respectively. The results clearly indicate that increasing the size of a language model does not significantly enhance detection performance, as zero-shot and few-shot learning did not yield notable improvements with larger models. However, the framework demonstrates that implementing a well-defined fine-tuning procedure, including partial fine-tuning by attaching adapters to a base language model, can yield higher accuracies, even with smaller models that lack the extensive knowledge base of their larger counterparts.

\section*{Ethical Consideration}
This paper proposes a framework for detecting implicit influential patterns embedded within conversations. No original data collection was conducted for this study; instead, we utilized publicly available datasets for our downstream tasks. Consequently, no approval for human subjects research was required. In the paper introducing the dataset \cite{wangMentalManipDatasetFinegrained2024}, the authors stated that the data was collected from movies and subsequently annotated by experts. For further details regarding the dataset’s data collection procedures, refer to the original publication by \cite{wangMentalManipDatasetFinegrained2024}.

\section*{Acknowledgement}
This work used the DeltaAI system at the National Center for Supercomputing Applications through allocation CIS250383 from the Advanced Cyberinfrastructure Coordination Ecosystem: Services \& Support (ACCESS) \cite{boernerACCESSAdvancingInnovation2023} program, which is supported by National Science Foundation grants \#2138259, \#2138286, \#2138307, \#2137603, and \#2138296.

\bibliographystyle{plain} 
\bibliography{Ref}
\section*{Appendices}
\subsection*{Appendix 1}\label{app:appendix1}
For data augmentation, conversations must be preprocessed to be suitable for prompting through a reasoning language model. First, line numbers were added before each sentence at the start of a conversation by each participant. This was necessary because, initially, the model was prompted to return sentences containing implicit influential patterns. Upon examining the model’s responses, we observed that sentences were sometimes only partially returned, and the structure of the answers was often inconsistent. Since our goal was to perform instruction fine-tuning, we assigned labels at the level of individual conversational turns, referred to as lines, rather than entire sentences. For example, regardless of whether a speaker's turn consisted of a single short sentence or multiple sentences, each turn was treated as a single line and assigned one label. For instance, in the example below, there are five conversational turns, and lines 1, 3 and 5 exhibit influential patterns. Thus, the correct label for instruction fine-tuning would be "Line\_1, Line\_3, Line\_5".

For the reasoning model, we selected DeepSeek-R1-Distill-Llama-8B. As this model supports "thinking tokens", we allowed it to process the conversation with intermediate reasoning steps, retrieving the final answer upon completion. Given the substantial number of tokens generated per prompt and the requirement for ten runs per conversation, the 8 billion parameters model was chosen for efficiency and computation constraints. The outputs from these ten runs were then aggregated and summarized using the Llama-3.3-70B-Instruct model. As the aggregation model only needed to process previously generated outputs rather than generating extensive new tokens, it proved even faster than the reasoning model during inference. The example below illustrates the data augmentation structure for a sample conversation.

\begin{longtable}{|p{3cm}|p{9cm}|}
\caption{An example of data augmentation prompts and their corresponding results} \label{tab3} \\
\hline
\textbf{Dialogue} & \textcolor{red}{Person1}: I like you so much. I think you're beautiful. I think if we were together you would love it. You wouldn't believe it. \newline \textcolor{blue}{Person2}: How do you know? \newline \textcolor{red}{Person1}: I just know. I know you'll love it. \newline \textcolor{blue}{Person2}: But I'm scared Telly. \newline \textcolor{red}{Person1}: I'm telling you. There's nothing in the world to worry about. \newline \textcolor{blue}{Person2}: Nothing? \\ \hline
\textbf{Dialogue with Labeled Lines} & \textcolor{JungleGreen}{Line\_1}: \textcolor{red}{Person1}: I like you so much. I think you're beautiful. I think if we were together you would love it. You wouldn't believe it. \newline \textcolor{JungleGreen}{Line\_2}: \textcolor{blue}{Person2}: How do you know? \newline \textcolor{JungleGreen}{Line\_3}: \textcolor{red}{Person1}: I just know. I know you'll love it. \newline \textcolor{JungleGreen}{Line\_4}: \textcolor{blue}{Person2}: But I'm scared Telly. \newline \textcolor{JungleGreen}{Line\_5}: \textcolor{red}{Person1}: I'm telling you. There's nothing in the world to worry about. \newline \textcolor{JungleGreen}{Line\_6}: \textcolor{blue}{Person2}: Nothing? \\ \hline
\textbf{Thinking Prompt} & In the provided conversation between two individuals, certain lines contain implicit manipulative remarks. Carefully read the entire conversation from beginning to end. Based on the full context, identify and extract the specific line(s) that contain implicit manipulative remarks. Do not evaluate lines in isolation — consider how each line fits within the overall dialogue. Return only the line numbers in the format 'Line\_X, Line\_Y, ...' without any explanation. For example, if the dialogue has lines A, B, C, D, E, and F and the implicit manipulation remarks occur in lines A and D, return 'Line\_A, Line\_D'. Implicit manipulation does not necessarily happen in multiple lines. It could be just happening in a single line. \\ \hline
\textbf{Full Prompt for Data Augmentation} & \textcolor{JungleGreen}{Line\_1}: \textcolor{red}{Person1}: I like you so much. I think you're beautiful. I think if we were together you would love it. You wouldn't believe it. \newline \textcolor{JungleGreen}{Line\_2}: \textcolor{blue}{Person2}: How do you know? \newline \textcolor{JungleGreen}{Line\_3}: \textcolor{red}{Person1}: I just know. I know you'll love it. \newline \textcolor{JungleGreen}{Line\_4}: \textcolor{blue}{Person2}: But I'm scared Telly. \newline \textcolor{JungleGreen}{Line\_5}: \textcolor{red}{Person1}: I'm telling you. There's nothing in the world to worry about. \newline \textcolor{JungleGreen}{Line\_6}: \textcolor{blue}{Person2}: Nothing? \newline \newline In the provided conversation between two individuals, certain lines contain implicit manipulative remarks. Carefully read the entire conversation from beginning to end. Based on the full context, identify and extract the specific line(s) that contain implicit manipulative remarks. Do not evaluate lines in isolation — consider how each line fits within the overall dialogue. Return only the line numbers in the format 'Line\_X, Line\_Y, ..
3 .' without any explanation. For example, if the dialogue has lines A, B, C, D, E, and F and the implicit manipulation remarks occur in lines A and D, return 'Line\_A, Line\_D'. Implicit manipulation does not necessarily happen in multiple lines. It could be just happening in a single line. \\ \hline
\textbf{Inference 1} & Line\_5 \\ \hline
\textbf{Inference 2} & Line\_1, Line\_5 \\ \hline
\textbf{Inference 3} & Line\_3, Line\_5 \\ \hline
\textbf{Inference 4} & Line\_1, Line\_3, Line\_5 \\ \hline
\textbf{Inference 5} & Line\_5 \\ \hline
\textbf{Inference 6} & Line\_3, Line\_5 \\ \hline
\textbf{Inference 7} & Line\_1, Line\_3, Line\_5 \\ \hline
\textbf{Inference 8} & Line\_1, Line\_3, Line\_5 \\ \hline
\textbf{Inference 9} & Line\_1, Line\_3, Line\_5 \\ \hline
\textbf{Inference 10} & Line\_1, Line\_3 \\ \hline
\textbf{Aggregation} & Line\_1, Line\_3, Line\_5 \\ \hline
\end{longtable}

\subsection*{Appendix 2}
The generative component of the proposed framework is illustrated with the following example. After the initial stage of instruction fine-tuning, the model can be prompted to identify influential segments within conversations. However, it is necessary to format the input as described in Appendix 1, with lines labeled accordingly, as this is the structure on which the model was trained. The example below demonstrates the generative aspect of the model:

\clearpage
\begin{table}[htbp]
\centering
\caption{An example demonstrating the generative capability of the proposed model}
\label{tab4}
\begin{tabularx}{\textwidth}{|p{3cm}|X|}
\hline
\textbf{Dialogue} & \textcolor{red}{Person1}: I don't know. \newline \textcolor{blue}{Person2}: Don't go. \newline \textcolor{red}{Person1}: I have to... he's a G-34 \newline \textcolor{blue}{Person2}: You don't have to... I don't trust him.  You ought to file a report. \newline \textcolor{red}{Person1}: No, don't make trouble.  It's nothing. \\ \hline
\textbf{Full prompt}      &
Analyze the following conversation for manipulative behavior: \newline \textcolor{JungleGreen}{Line\_1}: \textcolor{red}{Person1}: I don't know. \newline \textcolor{JungleGreen}{Line\_2}: \textcolor{blue}{Person2}: Don't go. \newline \textcolor{JungleGreen}{Line\_3}: \textcolor{red}{Person1}: I have to... he's a G-34 \newline \textcolor{JungleGreen}{Line\_4}: \textcolor{blue}{Person2}: You don't have to... I don't trust him.  You ought to file a report. \newline \textcolor{JungleGreen}{Line\_5}: \textcolor{red}{Person1}: No, don't make trouble.  It's nothing. \newline Identify the specific lines that demonstrate implicit manipulative intent.     \\ \hline
\textbf{Response}    & Line\_2, Line\_4 demonstrate implicit manipulative intent. Line\_2 is attempting to control Person1's actions by using a restrictive phrase ("Don't go"). Line\_4 is exerting influence over Person1's decision-making by expressing distrust and encouraging a report, which could be seen as an attempt to undermine Person1's autonomy.     \\ \hline
\end{tabularx}
\end{table}

\end{document}